%% file: root.tex
\documentclass[letterpaper, 10 pt, conference]{ieeeconf}  

\IEEEoverridecommandlockouts                              

\usepackage{graphicx}
\usepackage[flushleft]{threeparttable}
\usepackage{multirow}
\usepackage{xcolor}
\usepackage{amsmath}
\usepackage{amssymb}
\usepackage{cite}
\usepackage{times}
\usepackage{tabularx}
\usepackage{booktabs}
\usepackage{bm}
\usepackage{array}
\usepackage{soul}
\usepackage{algorithm}
\usepackage{algpseudocode}
\usepackage{colortbl}
\usepackage[colorlinks,linkcolor=red,anchorcolor=blue,citecolor=green]{hyperref}

\definecolor{darkgreen}{rgb}{0.0, 0.5, 0.0}

\newcommand{\ourmethod}{\textit{DreamDrive}}

\renewcommand{\baselinestretch}{0.985}

\title{\LARGE \bf
DreamDrive: Generative 4D Scene Modeling from Street View Images
}

\author{
        Jiageng Mao\textsuperscript{1,2}  Boyi Li\textsuperscript{1} Boris Ivanovic\textsuperscript{1} Yuxiao Chen\textsuperscript{1} Yan Wang\textsuperscript{1} Yurong You\textsuperscript{1} Chaowei Xiao\textsuperscript{1} Danfei Xu\textsuperscript{1}\\
        Marco Pavone\textsuperscript{1}  Yue Wang\textsuperscript{1,2} \\
        $^{1}$NVIDIA Research \quad $^{2}$University of Southern California
}

\begin{document}

\maketitle
\thispagestyle{empty}
\pagestyle{empty}

\begin{abstract}

Synthesizing photo-realistic visual observations from an ego vehicle's driving trajectory is a critical step towards scalable training of self-driving models. 
Reconstruction-based methods create 3D scenes from driving logs and synthesize geometry-consistent driving videos through neural rendering, but their dependence on costly object annotations limits their ability to generalize to in-the-wild driving scenarios. On the other hand, generative models can synthesize action-conditioned driving videos in a more generalizable way but often struggle with maintaining 3D visual consistency. In this paper, we present \ourmethod{}, a 4D spatial-temporal scene generation approach that combines the merits of generation and reconstruction, to synthesize generalizable 4D driving scenes and dynamic driving videos with 3D consistency. Specifically, we leverage the generative power of video diffusion models to synthesize a sequence of visual references and further elevate them to 4D with a novel hybrid Gaussian representation. Given a driving trajectory, we then render 3D-consistent driving videos via Gaussian splatting. 
The use of generative priors allows our method to produce high-quality 4D scenes from in-the-wild driving data, while neural rendering ensures 3D-consistent video generation from the 4D scenes.   
Extensive experiments on nuScenes and in-the-wild driving data demonstrate that \ourmethod{} can generate controllable and generalizable 4D driving scenes, synthesize novel views of driving videos with high fidelity and 3D consistency, decompose static and dynamic elements in a self-supervised manner, and enhance perception and planning tasks for autonomous driving. 




\end{abstract}

\input{secs/intro}
\input{secs/related}
\input{secs/method}

\input{secs/experiments}

\input{secs/conclusion}

\clearpage

\bibliographystyle{IEEEtran}
\bibliography{myref}

\end{document}

%% file: secs/intro.tex
\section{Introduction}



Generating driving videos based on ego vehicle's trajectory is a critical problem in autonomous driving. Action-conditioned video generation allows autonomous vehicles to anticipate future scenarios, respond accordingly, and generalize beyond expert trajectories, which is crucial for the scalable training of self-driving models.
To tackle this challenge, two series of works have emerged: reconstruction-based methods and generation-based methods. Reconstruction-based methods~\cite{nsg, unisim, emernerf, drivinggaussian, streetgaussian, s3gaussian, pvg} models 3D scenes from driving logs, and then action-conditioned visual observations can be generated through neural rendering techniques such as NeRF~\cite{nerf} or 3D Gaussian splatting~\cite{3dgs}. These methods can synthesize 3D-consistent and photo-realistic visual observations, However, they heavily rely on well-annotated driving logs that include calibrated camera poses, object boxes, and 3D point clouds, which limits their scalability to unlabeled in-the-wild driving data. On the other hand, generation-based methods~\cite{gaia1, magicdrive, drivewm, drivedreamer, genad, vista} can learn from in-the-wild driving data and synthesize action-conditioned dynamic driving videos via image~\cite{sd} or video diffusion models~\cite{svd}. However, video generation suffers from poor 3D geometry consistency across frames, which can undermine the reliability of the synthesized visual observations for autonomous driving. Hence, synthesizing both generalizable and 3D-consistent visual observations for autonomous driving remains an open challenge. 

\input{figs/teaser}


To address this challenge, we introduce \ourmethod{}, a 4D scene generation approach for autonomous driving. Our key insight is to combine the generative power of video diffusion priors with the geometry-consistent rendering of 3D Gaussian splatting~\cite{3dgs}. We elevate 2D visual references from video diffusion models into a 4D spatio-temporal scene, where the ego vehicle navigates and synthesizes novel-view observations through Gaussian splatting. Video diffusion priors enhance the generalization of our method, enabling 4D scene generation from in-the-wild driving data, while Gaussian splatting ensures 3D consistency during novel view synthesis. This approach allows \ourmethod{} to produce high-quality, 3D-consistent visual observations with strong generalization to diverse driving scenarios.

Although intuitive, accurately modeling 4D scenes from generated visual references remains quite challenging. Unlike well-annotated driving datasets~\cite{nuscenes, waymo}, generated visual references lack crucial information such as camera poses, object locations, and depth data, which hampers 4D modeling. Furthermore, this issue is compounded by the inherent 3D inconsistency in video diffusion models, causing traditional Gaussian representations~\cite{3dgs, deformablegs} to overfit to the training views and fail on novel view synthesis. To address these problems, we introduce a self-supervised hybrid Gaussian representation. 
Our approach leverages time-independent Gaussians to model static backgrounds and time-dependent Gaussians for dynamic objects, combining them into a unified 4D scene. First, we propose a self-supervised approach that can decompose a scene into static and dynamic regions with only image supervision. Next, we introduce spatio-temporal clustering to group 3D Gaussians into static and dynamic Gaussian clusters, which effectively mitigates fake dynamics in 4D modeling. Finally, we optimize the Gaussian clusters with time-dependent and time-independent representations to construct 4D scenes under image supervision. With the hybrid Gaussian representation, our method is able to synthesize 3D-consistent novel-view driving videos. Our method works with pure image supervision, eliminating the need for data annotations and making it more scalable and generalizable to in-the-wild driving data. 

We evaluate our approach on both the nuScenes dataset and in-the-wild driving scenarios, demonstrating the controllable and generalization capabilities of our 4D scene generation. Our method, using hybrid Gaussian representations, can generate high-quality, 3D-consistent novel-view driving videos, with a $30\%$ improvement in visual quality over previous methods~\cite{3dgs, deformablegs, magicdrive, streetgaussian, s3gaussian}. Additionally, we showcase the applications of our approach in perception and planning tasks for autonomous driving.



%% file: figs/teaser.tex
\begin{figure}[t]
    \centering
    \includegraphics[width=0.5\textwidth]{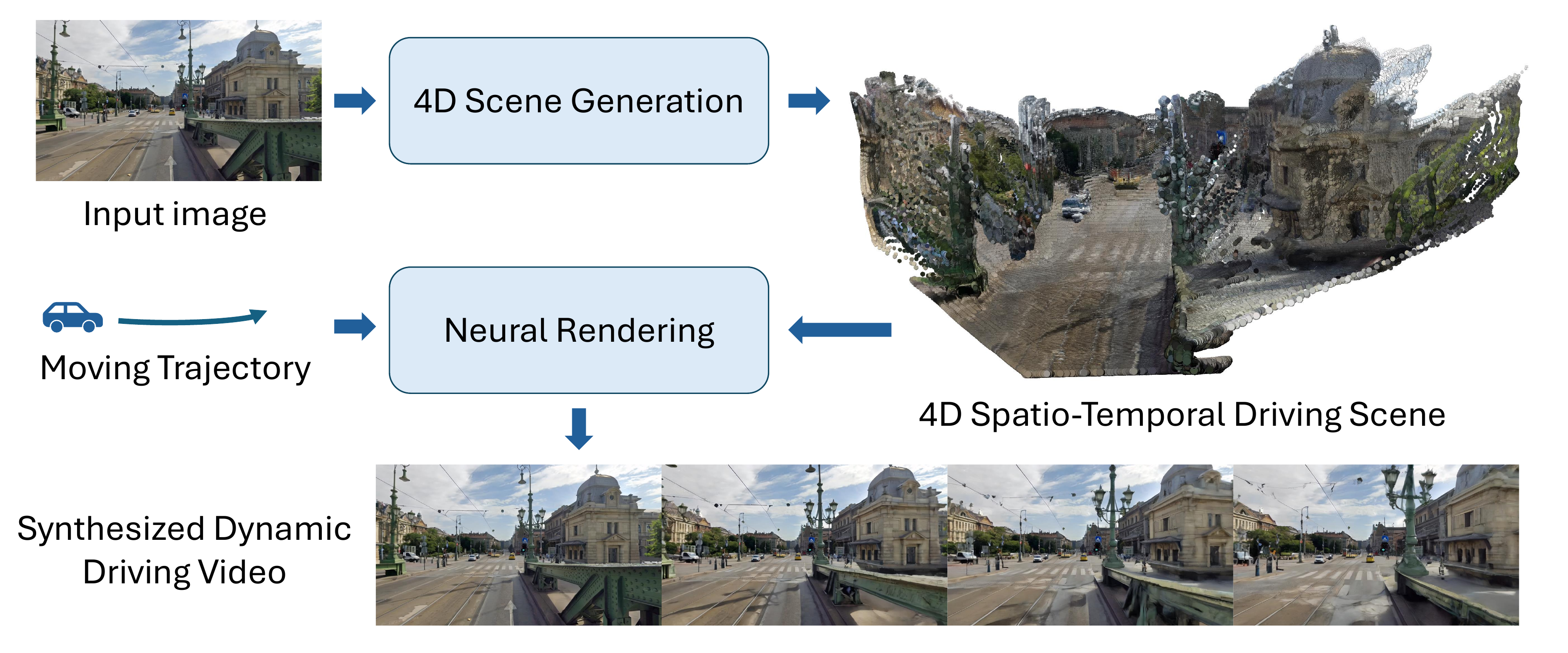}
    \vspace{-8mm}
    \caption{\textbf{An overview of \ourmethod.} Given an input image, our method can generate a 4D spatio-temporal driving scene, where we can render 3D-consistent dynamic driving videos with any driving trajectories.}
    \label{fig:teaser}
    \vspace{-5mm}
\end{figure}

%% file: secs/related.tex
\section{Related Works}

\textbf{Generative Models for Autonomous Driving.}
Generative models have shown significant potential in synthesizing future driving videos based on current actions. Recent works~\cite{drivedreamer, drivewm, bevgen, magicdrive, genad, vista, gaia1} have fine-tuned stable video diffusion models~\cite{svd} on driving data, incorporating controls such as map, object, weather, and action to generate diverse driving scenarios. However, as these models operate in 2D, they struggle to capture the underlying 3D geometry of the world, leading to poor 3D consistency in the generated videos. In contrast, our method employs neural rendering of 4D scenes, ensuring the generated videos maintain 3D consistency.


\textbf{Urban Scene Reconstruction.} Many papers~\cite{nsg, unisim, emernerf, mars, streetgaussian, drivinggaussian, pvg, s3gaussian} focus on reconstructing 3D or 4D urban scenes from driving logs, optimizing NeRF~\cite{nerf} or 3D-GS~\cite{3dgs} based scenes using multiview image supervision. These methods can synthesize novel views based on driving trajectories. However, most approaches~\cite{nsg, unisim, mars, streetgaussian, drivinggaussian} rely heavily on annotated object boxes to track and model dynamic objects, limiting their ability to handle unlabeled driving logs. While some methods~\cite{emernerf, s3gaussian, pvg} use self-supervised techniques to separate dynamic objects, they still depend on well-calibrated camera poses and 3D data, making them less generalizable to in-the-wild driving scenarios. In contrast, our method removes the need for pose or 3D information, enabling accurate 4D scene modeling directly from visual references.

\textbf{4D Scene Generation.} Many papers~\cite{sv4d, l4gm, dreamscene4d, diffusion4d, lgm, gaussianflow, comp4d, gslrm, 4k4dgen, textto4d, luciddreamer, alignyourgaussian, nfldm, lrm, grm, realmdreamer, one2345, zero123, dreamfusion, 4real, dreamgaussian4d, 4dfy, diffusionprior} focus on 3D and 4D content generation, but most focus on object generation, which is not applicable to driving scenarios. Some works~\cite{4real, 4dfy, dreamgaussian4d, dreamscene4d, diffusionprior} have introduced diffusion priors for 4D scene generation. However, the 4D scenes in these approaches are limited to object-centric, small-scale scenes, making it difficult for them to generalize to large-scale, unbounded driving scenes with numerous dynamic objects. 
The most relevant work,~\cite{magicdrive3d}, uses diffusion priors for 3D driving scene generation but relies solely on deformable 3D Gaussians, resulting in poor visual quality in novel view synthesis. 
In contrast, we propose a novel self-supervised approach to model 4D driving scenes with hybrid Gaussian representations, which demonstrates better generalization ability and visual quality in novel-view driving video synthesis.

%% file: secs/method.tex
\section{Method}

In this section, we introduce \ourmethod{}, a 4D spatio-temporal scene generation approach for autonomous driving. An overview of our method is shown in Figure~\ref{fig:method}. \ourmethod{} follows a 2D-3D-4D progressive generation process. We begin by leveraging \textit{video diffusion priors} to generate 2D visual references, followed by \textit{Gaussian initialization} to lift them into 3D. Next, we propose a novel \textit{self-supervised scene decomposition} approach with a \textit{clustering-based grouping} strategy to disentangle static and dynamic regions in the 4D spatio-temporal domain. Finally, we introduce \textit{hybrid Gaussian representations} to model static structures and dynamic objects for 4D scene generation. 

\input{figs/method}


\textbf{Problem Definition.} We study the problem of 4D scene generation. Given input controls, \textit{e.g.}, a single image $I_{ctrl}$ or a map with object locations $M_{ctrl}$, we aim to generate a 4D (3D+time) scene which is composed of a set of 3D Gaussians~\cite{3dgs}: $\{ G^{t}_{i} \mid i = 1, \dots, N_t, \ t = 1, \dots, T \}
$, where $N_t$ is the number of Gaussians at each timestep $t$, and $T$ is the total timesteps of this 4D scene. Each 3D Gaussian is parameterized by its mean position $x \in \mathbb{R}^{3}$, a quaternion-based rotation $r \in \mathbb{R}^{4}$ and scaling $s \in \mathbb{R}^{3}$, an opacity value $\alpha$, and a set of spherical harmonic (SH) coefficients $c$ to represent view-dependent color: $G(x, r, s, \alpha, c)$. The generation process can be formulated as
\begin{equation} \label{eq1}
    G = F_{gen}(X_{ctrl}), X_{ctrl} \in \{I_{ctrl}, M_{ctrl}\}, 
\end{equation}
where $F_{gen}$ is our proposed 4D scene generation approach. 

With this generated 4D scene representation, given any driving trajectory with camera poses $P_{traj} = \{P^{t} \mid t = 1, \dots, T \}$, we can synthesize a novel driving video $V = \{I^t \mid t = 1, \dots, T\}$ by splatting the 3D Gaussians $G^{t}$ at each timestep $t$ into an image $I^t$ with camera pose $P^t$:
\begin{equation} \label{eq2}
    I^t = F_{splat}(G^{t}, P^t),
\end{equation}
where $F_{splat}$ is the 3D Gaussian splatting in~\cite{3dgs}. Our method generates 4D driving scenes with diverse controls $X_{ctrl}$, and the neural rendering function $F_{splat}$ ensures the spatio-temporal consistency of synthesized driving videos.

\textbf{Video Diffusion Priors.} Video diffusion models are highly effective at modeling the temporal dynamics of visual data, but relying solely on them for trajectory-conditioned video generation can lead to 3D inconsistency, as they are designed for 2D image generation without considering the underlying 3D structure. In our method, we use video diffusion priors to generate initial visual references, which are then elevated to the 4D space for scene generation and 3D-consistent video rendering. Specifically, we employ video diffusion models~\cite{vista, magicdrive} trained on driving data to generate a sequence of reference images $\{I^{t}_{ref} \mid t = 1, \dots, T\}$ and extract latent features $Z_{ref}$ from the early layers to capture valuable visual dynamics for static-dynamic decomposition.
The process is formally expressed as:
\begin{equation} \label{eq3}
    I_{ref}, Z_{ref} = F_{VDM}(X_{ctrl}),
\end{equation}
where $F_{VDM}$ is the video diffusion model and $X_{ctrl}$ is the input control. $F_{VDM}$ provides visual references that guide 4D scene generation. Since it can generate references from in-the-wild driving data, incorporating video diffusion priors improves the generalization of our approach.

\textbf{Gaussian Initialization.}
Lifting generated images ${I_{ref}}$ into 4D space is quite challenging without camera poses and 3D information. Therefore, robust estimation of both camera parameters and 3D structure is crucial as a reliable initialization for 4D scene generation. 
While previous works use COLMAP~\cite{colmap} to estimate coarse 3D geometry, its sparse point clouds are insufficient for modeling large-scale and unbounded driving scenes. Instead, we employ an end-to-end multiview stereo network~\cite{dust3r} to produce pixel-aligned dense 3D geometry and simultaneously recover camera poses $\{P^t_{ref} \mid t = 1, \dots, T\}$. 
Specifically, \cite{dust3r} generates dense, pixel-aligned 3D point clouds for each image. Camera intrinsics are estimated using the Weiszfeld algorithm~\cite{weiszfeld}, and camera extrinsics are computed by globally aligning the point clouds across frames. The aggregated point clouds form a dense scene-level point cloud, which is used to initialize 3D Gaussian parameters, yielding a set of Gaussians $G_{init}$. These 3D Gaussians are further enriched with pixel-aligned latent features $Z_{ref}$. The whole process can be expressed as:
\begin{equation} \label{eq4}
   G_{init}, P_{ref} = F_{MVS}(I_{ref}, Z_{ref}),
\end{equation}
where $F_{MVS}$ is the multiview stereo network. This approach ensures accurate 3D scene geometry and camera estimation and serves as a robust initialization of 3D Gaussians.

With the initialized 3D Gaussians $G_{init}$, the next step is to model 4D spatio-temporal driving scenes containing both static backgrounds and dynamic objects. Previous works~\cite{emernerf, unisim, streetgaussian, drivinggaussian} rely on annotated object boxes to track dynamic objects, limiting their generalization to unannotated data like $I_{ref}$. Other methods~\cite{s3gaussian, pvg, magicdrive3d} use pure time-dependent Gaussians that change positions and shapes over time, but the 3D inconsistency in generated images often leads to overfitting and introduces fake dynamics, such as visual deformation in static structures when synthesizing novel views. To overcome these issues, we propose a novel hybrid Gaussian representation to model static and dynamic components separately. We divide the initial Gaussians $G_{init}$ into time-independent static Gaussians $G_{static}$ and time-dependent dynamic Gaussians $G_{dynamic}$, effectively modeling static structures and dynamic objects. This separation ensures that static structures remain consistent over time, mitigating fake dynamics while accurately capturing the movement of dynamic objects.

\textbf{Self-Supervised Scene Decomposition.} A key challenge in hybrid modeling is separating static and dynamic regions without additional annotations. To tackle this, our key insight is that image error maps serve as effective indicators for distinguishing between static and dynamic regions. Specifically, we first optimize the entire scene by assuming all initial Gaussians $G_{init}$ are \textit{static}. We then splat the optimized static Gaussians into static images $I_{static}$:
\begin{equation} \label{eq5}
    I^t_{static} = F_{splat}(G_{init}, P^t_{ref}).    
\end{equation}
Next, the error map at each timestep $t$ is computed as: 
\begin{equation} \label{eq6}
    I^t_{err} = |I^t_{static} - I^t_{ref}|.
\end{equation}
The pixels in $I_{err}$ with higher rendering errors indicate the regions that static Gaussians struggle to optimize, suggesting that these areas likely correspond to dynamic objects. Therefore, we can use $I_{err}$ as supervisory signals for scene decomposition. In particular, we train a network, $F_{score}$, that takes the initial Gaussians ${G_{init}}$ and their associated latent features ${Z_{ref}}$ as input, and outputs binary dynamic scores $S$ to classify each Gaussian as static or dynamic:
\begin{equation} \label{eq7}
    S = F_{score}(G_{init}, Z_{ref}),
\end{equation}
These scores are splatted into image planes using the Gaussian splatting function $F_{splat}$, and supervised with error maps $I_{err}$ using the binary cross-entropy loss $L_{bce}$:
\begin{equation} \label{eq8}
    L_{dec} = \sum_{t=0}^T (L_{bce}(F_{splat}(S, P^t_{ref}), I^t_{err})).
\end{equation}
Since the splatting function $F_{splat}$ is differentiable, we can optimize the scoring network $F_{score}$ end-to-end using the image-based decomposition loss $L_{dec}$. Finally, we separate the initial Gaussians $G_{init}$ into static Gaussians $G^{\prime}_{static}$ and dynamic Gaussians $G^{\prime}_{dyn}$ by applying a threshold $\tau$ to the predicted dynamic scores $S$:
\begin{equation} \label{eq9}
    G^{\prime}_{dynamic} = \{G_{init} \mid S > \tau \}, G^{\prime}_{static} = \{G_{init} \mid S \leq \tau \}.
\end{equation}
Unlike previous methods~\cite{dnerf, d2nerf, nerfhugs, nerfonthego, robustnerf, wildgaussians, wildgs, spotlesssplats}, our self-supervised approach doesn't require annotations or multiple passes, making it more scalable for large-scale driving scenes.

\textbf{Grouping with Gaussian Clusters.} Due to the inherent 3D inconsistencies in generated visual references, fake dynamics, such as local deformations in static structures, often appear in $I_{ref}$. This results in the incorrect assignment of dynamic Gaussians to static objects and negatively impacts 4D scene modeling and novel view synthesis. To improve the robustness of our scene decomposition, we introduce a novel cluster-based grouping strategy. Our key insight is that objects generally move as a whole, \textit{i.e.} Gaussians in the same object are likely to have the same dynamic attribute. As we don't have object annotations, we instead introduce ``spatio-temporal clustering" to group the Gaussians into clusters. If most Gaussians in a cluster are static, meaning that the whole part should be static, we assign static labels to all, even if some were initially classified as dynamic, and vice versa for dynamic clusters. The process can be expressed as
\begin{equation} \label{eq10}
    G_{static}, G_{dynamic} = F_{group}(G^{\prime}_{static}, G^{\prime}_{dynamic}),
\end{equation}
where $F_{group}$ is the proposed grouping strategy. $F_{group}$ helps to rectify incorrect dynamic score predictions. We find this can efficiently reduce fake dynamics, leading to more accurate and consistent 4D scene modeling.

\textbf{Hybrid Gaussian Representations.} Scene decomposition enables us to represent static and dynamic components with distinct Gaussians. Static Gaussians $G_{static}$ model elements like roads and buildings, with parameters $G(x, r, s, \alpha, c)$ that remain constant over time, ensuring accurate rendering of static structures. Dynamic Gaussians $G_{dynamic}$ model objects like cars and pedestrians, where Gaussian positions and shapes vary over time: $G^t_{dynamic} = G(x^t, r^t, s^t, \alpha, c)$. We follow~\cite{deformablegs} and learn a deformation network $F_{deform}$ that takes the Gaussian positions $x$ and a timestep $t$ as input and predicts temporal offsets of the Gaussians: $(\delta x, \delta r, \delta s)$:
\begin{equation} \label{eq11}
    (\delta x, \delta r, \delta s) = F_{deform}(x, t),
\end{equation}
\begin{equation} \label{eq12}
    (x^t, r^t, s^t, \alpha, c) = (x + \delta x, r + \delta r, s + \delta s, \alpha, c).
\end{equation}
These time-dependent dynamic Gaussians $G_{dynamic}$ accurately represent dynamic objects in 4D scenes.

Finally, we combine $G_{static}$ and $G_{dynamic}$ into a 4D spatio-temporal scene and optimize their parameters by splatting them onto images $I^t_{render}$ at each timestep $t$: 
\begin{equation} \label{eq13}
    I^t_{render} = F_{splat}(\{G_{static}, G^t_{dynamic}\}, P^t_{ref}).
\end{equation}
The rendering loss can be computed as:
\begin{equation} \label{eq14}
    L_{render} = \sum^T_{t=0} (L_{1}(I^t_{render}, I^t_{ref}) + L_{SSIM}(I^t_{render}, I^t_{ref})),
\end{equation}
where $L_{SSIM}$ is the SSIM loss~\cite{ssimloss}. We jointly optimize Gaussian parameters and $F_{deform}$ based on the rendering loss $L_{render}$, leading to robust 4D scene modeling.

%% file: figs/method.tex
\begin{figure*}[t]
    \centering
    \includegraphics[width=\textwidth]{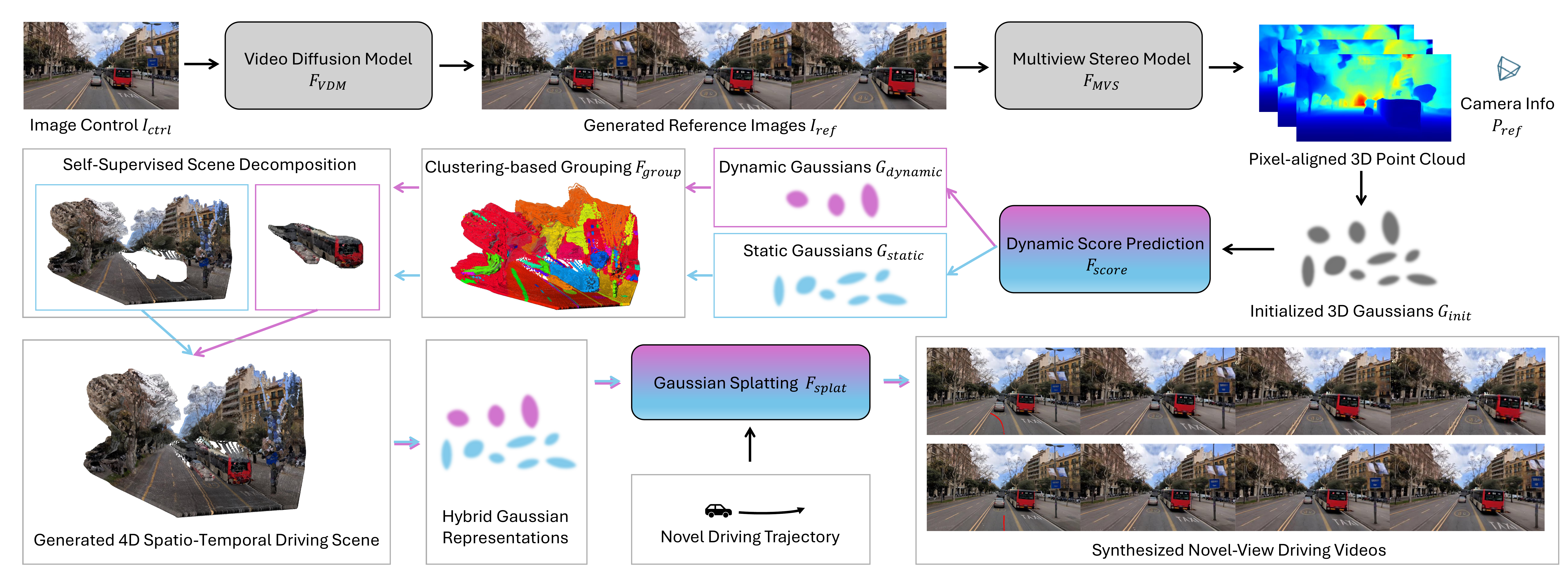}
    \vspace{-8mm}
    \caption{\textbf{\ourmethod{} model pipeline.} Given an input control image $I_{ctrl}$, our method first generates a set of reference images $I_{ref}$ using a video diffusion model $F_{VDM}$. These reference images are then lifted into 3D space via a multiview stereo network $F_{MVS}$, which provides camera information and dense 3D scene geometry to initialize 3D Gaussians. Next, we employ a self-supervised scoring network $F_{score}$ to separate the 3D Gaussians into static and dynamic components, followed by a clustering-based grouping strategy that creates hybrid Gaussian representations for modeling static structures and dynamic objects in the 4D spatio-temporal driving scene. Finally, we optimize the 4D scene using supervision from the reference images. During inference, given a driving trajectory, a novel-view driving video can be synthesized by splatting the hybrid Gaussian representations into images at each timestep.}
    \label{fig:method}
    \vspace{-5mm}
\end{figure*}

%% file: secs/experiments.tex
\input{figs/control_demo}

\section{Experiments}

This section investigates the following questions:

$\bullet$ Can \ourmethod{} generate 4D driving scenes in a controllable and generalizable way? (Section \ref{E2})

$\bullet$ Can \ourmethod{} synthesize novel-view driving videos with high fidelity and 3D consistency? (Section \ref{E3})

$\bullet$ Can \ourmethod{} decompose static background and dynamic objects in a self-supervised manner? How does the decomposition help 4D scene modeling and novel-view driving video synthesis? (Section \ref{E4})

$\bullet$ Can \ourmethod{} help onboard autonomous driving tasks such as perception and planning? (Section \ref{E5})

\subsection{Experimental Setup} \label{E1}

We utilize two complementary datasets to assess the performance of our method across both controlled and in-the-wild driving scenarios, each presenting distinct challenges. For controlled driving scenarios, we use the nuScenes dataset~\cite{nuscenes}, a large-scale real-world autonomous driving dataset comprising 1,000 driving sequences and approximately 34,000 key frames, all with accurately calibrated camera poses, maps, and object annotations. This makes it an ideal choice for validating our method in controlled settings. Following standard practice, we divide the dataset into training and validation sets. For in-the-wild driving scenarios, we curate 20 scenarios from various geographical regions using Google Street View. This benchmark allows us to evaluate the generalization ability of our approach. We leverage the video diffusion prior from \cite{magicdrive} for the nuScenes benchmark and from \cite{vista} for the in-the-wild benchmark.

\subsection{Controllable and Generalizable 4D Scene Generation} \label{E2}

\input{figs/generalize_demo}

\input{figs/traj_demo}

In Figure~\ref{fig:control_demo}, we demonstrate the controllability of \ourmethod{} in generating 3D scenes. Our approach allows for fine-grained control over scene elements such as road layouts and object positions. This controllability stems from the synergy of generative models and reconstruction methods, providing the ability to manipulate individual elements not only in images but also in 3D while maintaining high fidelity.
Figure~\ref{fig:generalize_demo} further illustrates \ourmethod{}’s generalization ability. By using only in-the-wild images, \ourmethod{} successfully generates realistic 4D scenes from diverse geographical locations such as Japan, Australia, and the United States. Unlike traditional approaches that rely heavily on labeled datasets or precise calibration data, our method employs self-supervised learning to model 4D driving scenes without the need for exhaustive manual annotations. This adaptability allows the method to work across various sensory setups and eliminates the requirement for specialized data collection, demonstrating its robustness in diverse driving scenarios. 


\subsection{Novel-View Video Synthesis with 3D Consistency} \label{E3}

\input{figs/videogen}

\input{tables/fid}

Given a driving trajectory, \ourmethod{} synthesizes dynamic driving videos by rendering the generated 4D scenes into images. Figure~\ref{fig:traj_demo} presents examples of novel-view driving videos generated by our approach, covering various driving maneuvers such as moving forward, turning left, and stopping. Unlike previous methods that struggle with geometric consistency when changing viewpoints, \ourmethod{} maintains spatial accuracy for static and dynamic elements, ensuring realistic and consistent driving video generation. 
Furthermore, as shown in Figure~\ref{fig:videogen}, compared to directly generating videos with diffusion models, \ourmethod{} offers more precise trajectory control and better 3D consistency by leveraging 4D scene generation and neural rendering. 
As shown in Table~\ref{tab:fid}, our method achieves the lowest FID of 45.59 and FVD of 374.02, significantly improving over previous methods like self-supervised street Gaussians~\cite{s3gaussian} and MagicDrive3D~\cite{magicdrive3d}. These advancements are due to our self-supervised decomposition module, which accurately separates static backgrounds from dynamic objects for more precise scene representation.

\subsection{Static-Dynamic Decomposition for 4D Scenes} \label{E4}


The static-dynamic decomposition is crucial for effective 4D scene modeling in \ourmethod{}. Unlike prior methods that rely on annotated object boxes~\cite{streetgaussian, drivinggaussian} or treat the entire scene as dynamic~\cite{s3gaussian, deformablegs, magicdrive3d}, \ourmethod{} uses a self-supervised approach to segment moving objects from static environments (Figure~\ref{fig:dynamic}), eliminating the need for costly annotations and improving scalability in diverse driving scenarios. As shown in Figure~\ref{fig:deform}, deformable Gaussians~\cite{magicdrive3d, deformablegs} often overfit to training views, producing poor results in novel-view synthesis. In contrast, \ourmethod{} employs a hybrid Gaussian representation, \textit{i.e.}, time-independent Gaussians for static backgrounds and time-dependent Gaussians for dynamic objects, which improves robustness and accuracy in motion capture, reduces fake dynamics in static background, and ensures consistent 4D scene modeling.

\input{figs/dynamic}

\input{figs/deform}

\subsection{Training Support for Perception and Planning} \label{E5}

\input{tables/perception}

\ourmethod{} improves the perception and planning capabilities of autonomous vehicles in multiple ways. For perception, it generates diverse 3D scenes from map layouts and object locations, and through neural rendering, synthesizes view-consistent images that serve as training data. To evaluate this, we generate 3D scenes from the nuScenes training set and use the synthetic images to train BEV segmentation models~\cite{cvt}. We evaluate our method by training the model on both synthetic and real data, to see if there is a performance improvement, as well as training on synthetic data alone, to see if it can match real data performance. As shown in Table~\ref{tab:perception}, using \ourmethod{} as data augmentation significantly improves BEV segmentation achieving 37.19 vehicle mIOU and 73.03 road mIOU. Even when trained solely on our generated data, \ourmethod{} achieves higher performance than baseline methods~\cite{bevcontrol, bevgen}, demonstrating the quality of our generated data.


\ourmethod{} can also help planning in autonomous driving. Neural motion planners can be trained on synthetic data, and since our method generates 4D scenes, we can further optimize the planning trajectories by checking their collisions with 3D Gaussians in the spatio-temporal domain (Figure~\ref{fig:planning}). To validate this, we optimize the planned trajectories in~\cite{gptdriver} by minimizing the cost functions in \cite{uniad}, treating filtered 3D Gaussians as occupied points. Results in Table~\ref{tab:planning} demonstrate that our method could reduce the collision rate by $25\%$.

\input{tables/planning}

\input{figs/planning}

%% file: figs/control_demo.tex
\begin{figure*}[htp]
    \centering
    \includegraphics[width=\textwidth]{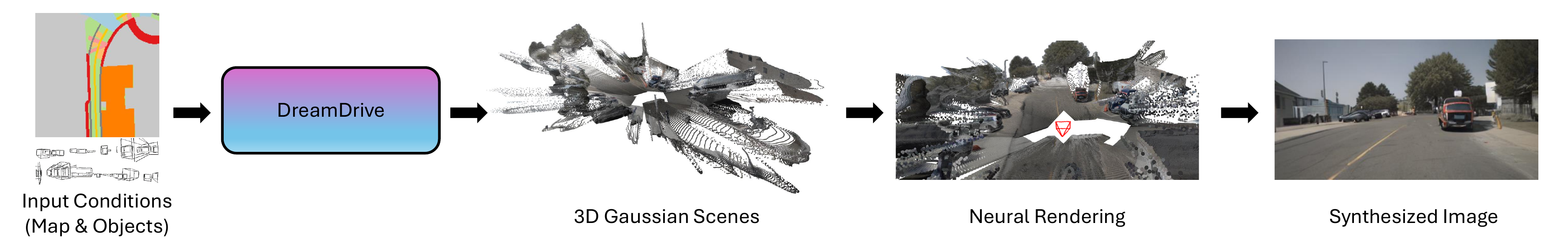}
    \vspace{-9mm}
    \caption{\textbf{Controllability of \ourmethod}. Our method generates 3D Gaussian scenes with map and object control.}
    \label{fig:control_demo}
    \vspace{-4mm}
\end{figure*}

%% file: figs/generalize_demo.tex
\begin{figure*}[htp]
    \centering
    \includegraphics[width=\textwidth]{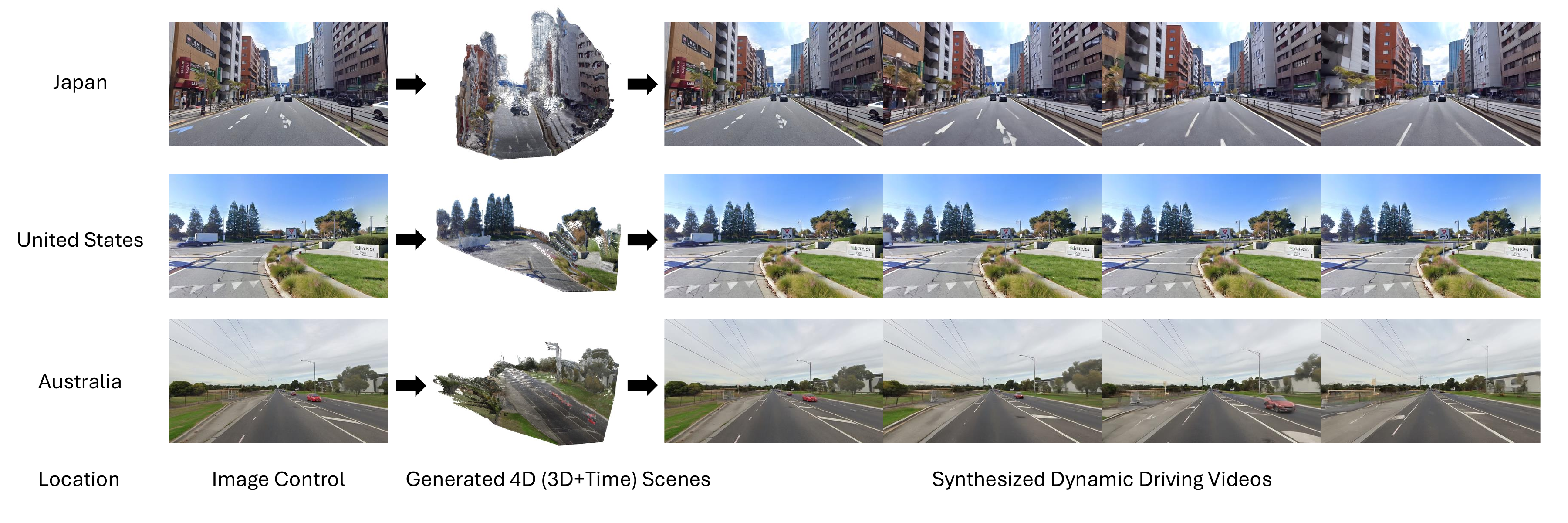}
    \vspace{-10mm}
    \caption{\textbf{Generalization ability of \ourmethod{}.} Given an image from anywhere in the world, our method can generate a 4D scene and render 3D-consistent driving videos from the 4D scene. This eliminates the requirement for specialized data collection and enables us to drive everywhere in the 3D world.}
    \label{fig:generalize_demo}
    \vspace{-3mm}
\end{figure*}

%% file: figs/traj_demo.tex
\begin{figure*}[htp]
    \centering
    \includegraphics[width=\textwidth]{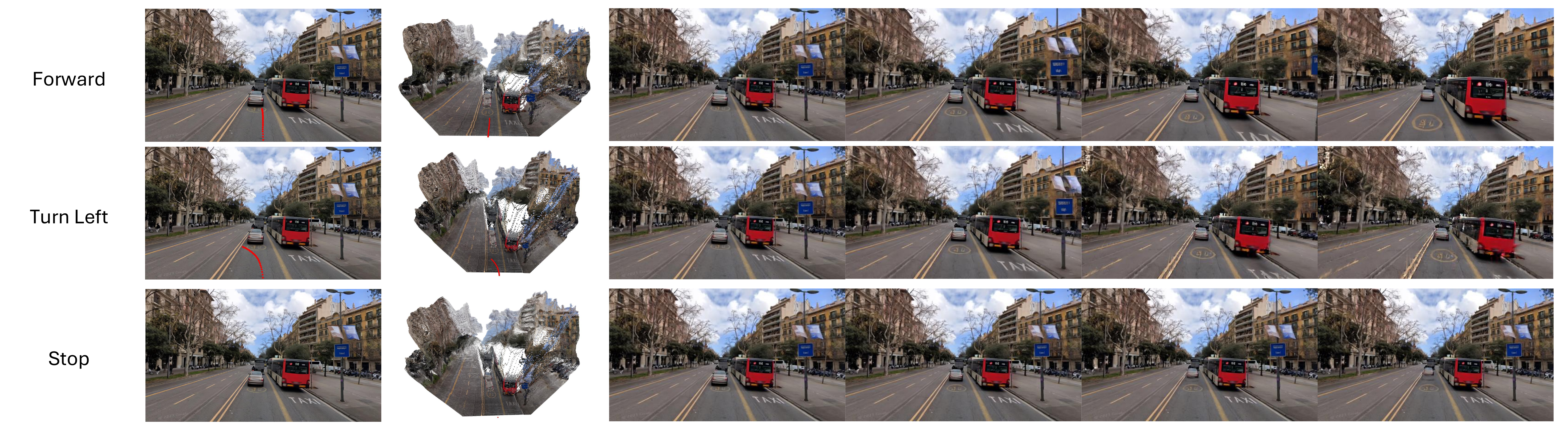}
    \vspace{-7mm}
    \caption{\textbf{Novel-view driving video synthesis in \ourmethod{}.} Our method can generate geometry-consistent driving videos with different driving trajectories.}
    \label{fig:traj_demo}
    \vspace{-4mm}
\end{figure*}

%% file: figs/videogen.tex
\begin{figure}[htp]
    \centering
    \includegraphics[width=0.5\textwidth]{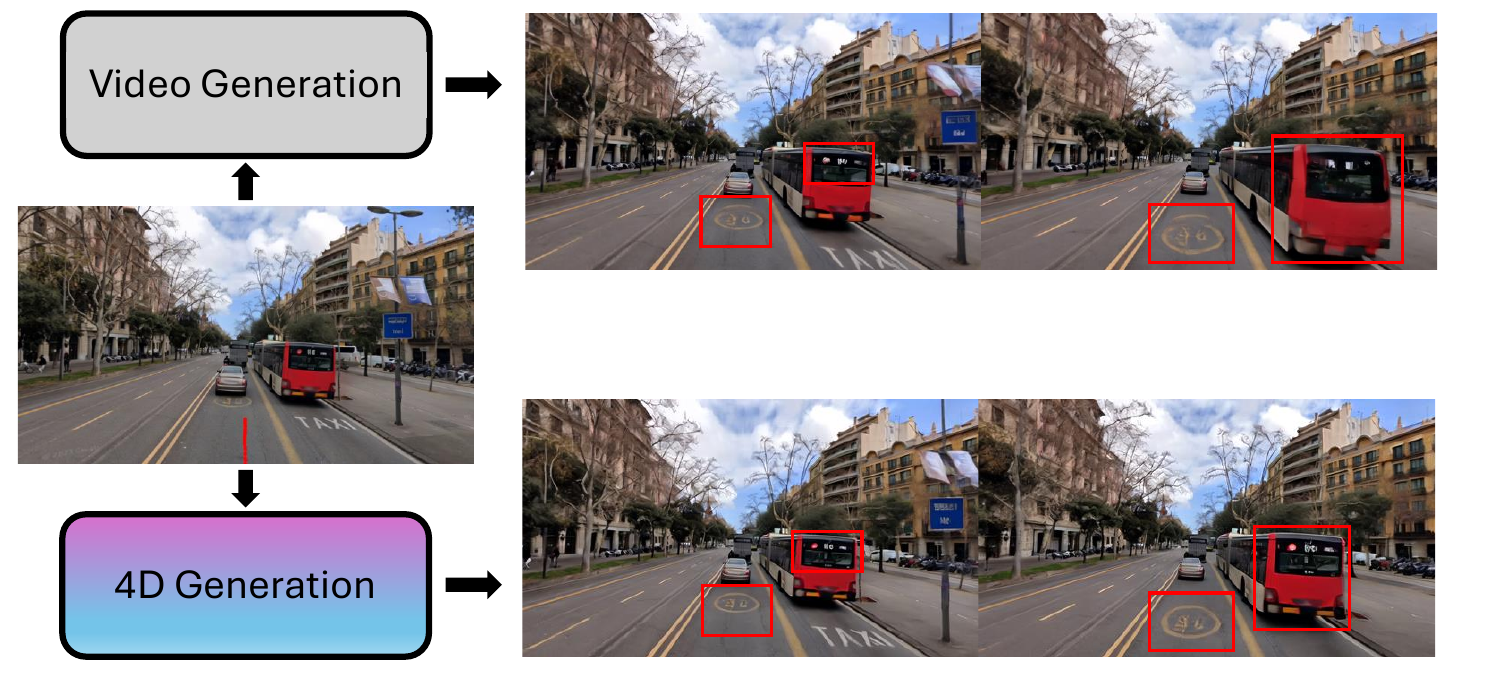}
    \vspace{-8mm}
    \caption{\textbf{Video generation vs. 4D scene generation \& neural rendering.} Compared to directly generating videos with diffusion models, our method offers more precise trajectory controls and better 3D consistency.}
    \label{fig:videogen}
    \vspace{-3mm}
\end{figure}

%% file: tables/fid.tex
\begin{table}[th]
  \renewcommand{\arraystretch}{1.1}
  \renewcommand\tabcolsep{12.5pt} 
  \footnotesize
  \caption{Novel-view video generation quality of DreamDrive.}
    \vspace{-5mm}
    \begin{center}
    \begin{tabular}{l|cc}
        \toprule[.03cm]
        Method & FID $\downarrow$ & FVD $\downarrow$ \\
        \midrule
        3D-GS~\cite{3dgs} & 82.12 & 787.67 \\
        Street Gaussians~\cite{s3gaussian, streetgaussian} & 61.19 & 456.56 \\
        Deformable Gaussians~\cite{deformablegs, magicdrive3d} & 65.29 & 450.43 \\
        \ourmethod\space  (Ours) & \textbf{45.59} & \textbf{374.02}  \\
        
        \toprule[.03cm]
    \end{tabular}
    \end{center} 
    \label{tab:fid}
    \vspace{-10mm}
\end{table}


%% file: figs/dynamic.tex
\begin{figure}[thp]
    \centering
    \includegraphics[width=0.5\textwidth]{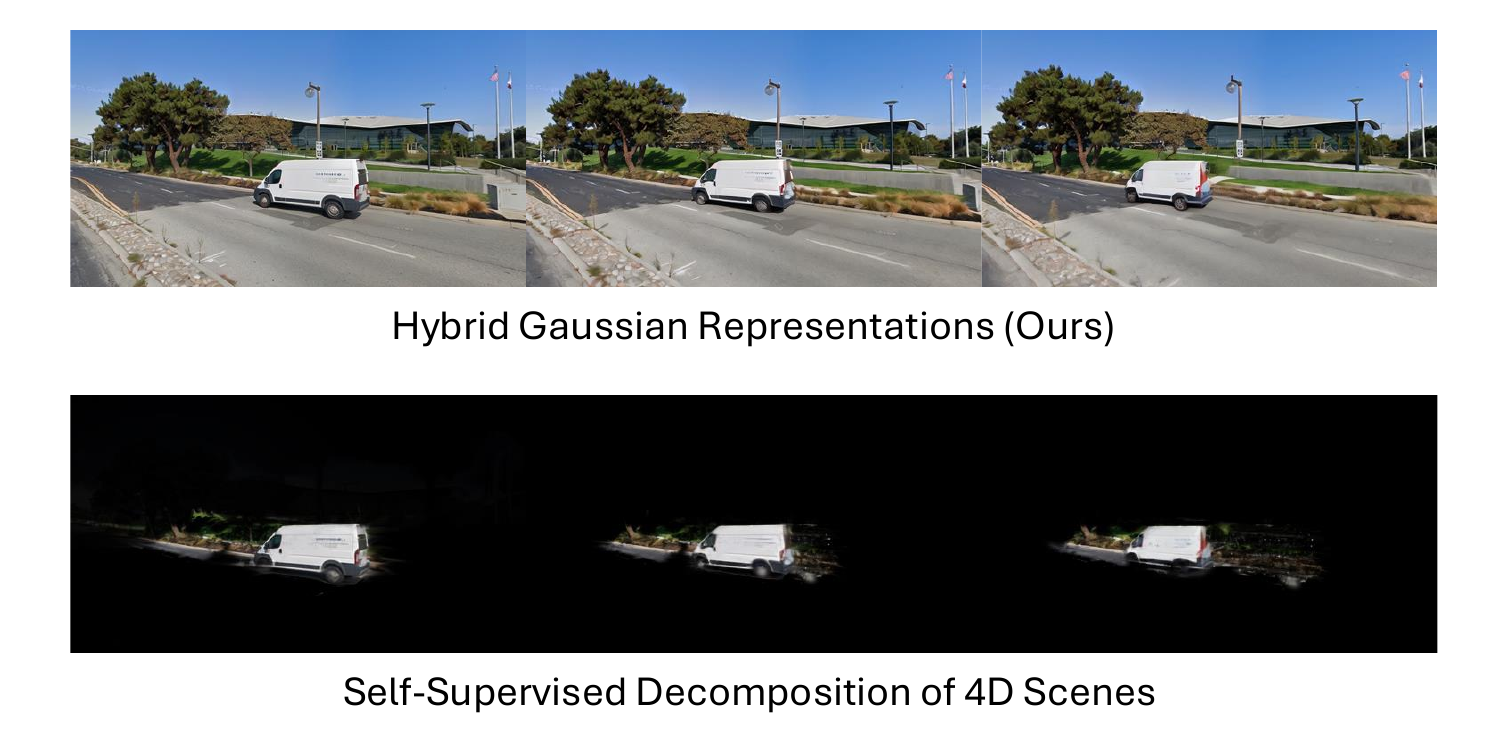}
    \vspace{-9mm}
    \caption{\textbf{Effectiveness of 4D scene decomposition in \ourmethod.}}
    \label{fig:dynamic}
    \vspace{-5mm}
\end{figure}

%% file: figs/deform.tex
\begin{figure}[htp]
    \centering
    \includegraphics[width=0.5\textwidth]{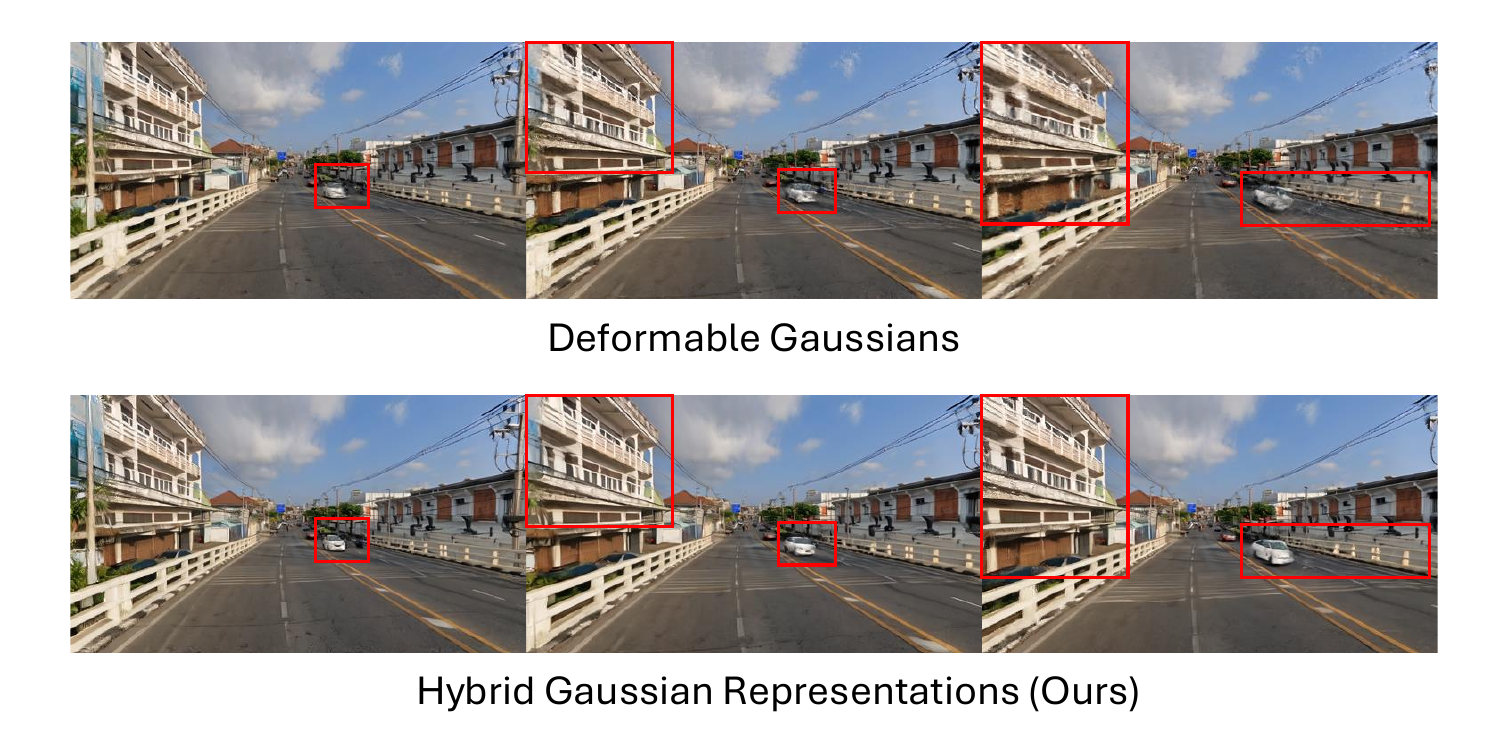}
    \vspace{-9mm}
    \caption{\textbf{Effectiveness of hybrid Gaussian representations in \ourmethod.}}
    \label{fig:deform}
    \vspace{-3mm}
\end{figure}

%% file: tables/perception.tex
\begin{table}[th]
  \renewcommand{\arraystretch}{1.1}
  \renewcommand\tabcolsep{4.5pt} 
  \footnotesize
  \caption{Training support of DreamDrive for perception task.}
    \vspace{-8mm}
    \begin{center}
    \begin{tabular}{cc|c|c}
        \toprule[.03cm]
        \multicolumn{2}{c|}{\multirow{2}{*}{Method}}   & \multicolumn{2}{c}{{BEV Segmentation}} \\
        \cline{3-4} \rule{0pt}{10pt}
         &  & Vehicle mIOU $\uparrow$ & Road mIOU $\uparrow$  \\
        \midrule
        \multirow{3}*{\centering\shortstack{As Data\\Augmentation}} & \multicolumn{1}{|l|}{Oracle~\cite{cvt}} & 36.00 & 71.68  \\
        & \multicolumn{1}{|l|}{w/ BEV Gen~\cite{bevgen}} & 36.60 & 71.90 \\
        & \multicolumn{1}{|l|}{w/ \ourmethod} & \textbf{37.19}  & \textbf{73.03}  \\
        \midrule
        \multirow{3}*{\centering\shortstack{Synthetic Data\\Only}} & \multicolumn{1}{|l|}{Oracle~\cite{cvt}} & 36.00 & 71.68  \\
        & \multicolumn{1}{|l|}{BEV Gen~\cite{bevgen}} & 5.89 & 50.20 \\
        & \multicolumn{1}{|l|}{BEV Control~\cite{bevcontrol}} & 26.80 & 60.80 \\
        & \multicolumn{1}{|l|}{\ourmethod} & \textbf{32.62} & \textbf{70.83}  \\
        \toprule[.03cm]
    \end{tabular}
    \end{center} 
    \label{tab:perception}
    \vspace{-6mm}
\end{table}


%% file: tables/planning.tex

\begin{table}[ht]
\footnotesize
\caption{Trajectory optimization of DreamDrive for motion planning.}
\vspace{-7mm}
\begin{center}
\resizebox{0.49\textwidth}{!}{
\begin{tabular}{l|cccc|cccc}
\toprule

\multirow{2}{*}{Method} &
\multicolumn{4}{c|}{L2 (m) $\downarrow$} & 
\multicolumn{4}{c}{Collision (\%) $\downarrow$} \\
 & 1s & 2s & 3s & \cellcolor{gray!30}Avg. & 1s & 2s & 3s & \cellcolor{gray!30}Avg. \\

\midrule
\multicolumn{9}{c}{ST-P3 metrics} \\
\midrule
ST-P3~\cite{stp3} & 1.33 & 2.11 & 2.90 & \cellcolor{gray!30}2.11 & 0.23 & 0.62 & 1.27 & \cellcolor{gray!30}0.71 \\
VAD~\cite{vad} & 0.17 & 0.34 & 0.60 & \cellcolor{gray!30}0.37 & 0.07 & 0.10 & 0.24 & \cellcolor{gray!30}0.14 \\
\cmidrule(){1-9}
GPT-Driver~\cite{gptdriver} & 0.20 & 0.40 & 0.70 & \cellcolor{gray!30}0.44 & 0.04 & 0.12  & 0.36 & 0.17\cellcolor{gray!30} \\
w/ DreamDrive & 0.21 & 0.41 & 0.72 & \cellcolor{gray!30}0.45 & 0.03 & 0.08  & 0.30 & \textbf{0.14}\cellcolor{gray!30} \\
\midrule
\multicolumn{9}{c}{UniAD metrics} \\
\midrule
NMP~\cite{nmp} & - & - & 2.31 & \cellcolor{gray!30}- & - & - & 1.92 & \cellcolor{gray!30}- \\
SA-NMP~\cite{nmp} & - & - & 2.05 & \cellcolor{gray!30}- & - & - & 1.59 & \cellcolor{gray!30}- \\
FF~\cite{ff} & 0.55 & 1.20 & 2.54 & \cellcolor{gray!30}1.43 & 0.06 & 0.17 & 1.07 & \cellcolor{gray!30}0.43 \\
EO~\cite{eo} & 0.67 & 1.36 & 2.78 & \cellcolor{gray!30}1.60 & 0.04 & 0.09 & 0.88 & \cellcolor{gray!30}0.33 \\
UniAD~\cite{uniad} & 0.48 & 0.96 & 1.65 & \cellcolor{gray!30}1.03 & 0.05 & 0.17 & 0.71 & \cellcolor{gray!30}0.31 \\
\cmidrule(){1-9}
GPT-Driver~\cite{gptdriver} & 0.27  & 0.74 & 1.52 & \cellcolor{gray!30}0.84 & 0.07 & 0.15 & 1.10 & \cellcolor{gray!30}0.44 \\
w/ DreamDrive  & 0.28  & 0.76 & 1.56 & \cellcolor{gray!30}0.87 & 0.02 & 0.13 & 0.76 & \cellcolor{gray!30}\textbf{0.30} \\
\bottomrule
\end{tabular} }
\end{center}
\label{tab:planning}
\vspace{-8mm}
\end{table}

%% file: figs/planning.tex
\begin{figure}[htp]
    \centering
    \includegraphics[width=0.5\textwidth]{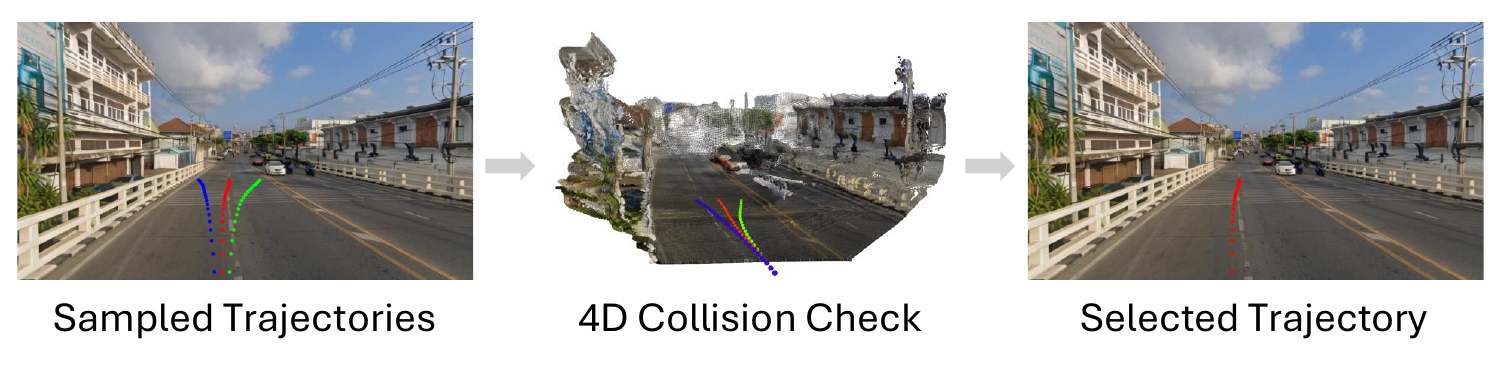}
    \vspace{-8mm}
    \caption{\textbf{Demonstration of how \ourmethod{} helps motion planning.}}
    \label{fig:planning}
    \vspace{-3mm}
\end{figure}

%% file: secs/conclusion.tex
\section{Conclusion}

In this paper, we present \ourmethod{}, a novel 4D scene generation approach for autonomous driving that combines the generative power of video diffusion models with the geometric consistency of 3D Gaussian splatting. Using hybrid Gaussian representations, our method accurately models both static and dynamic elements in 4D driving scenes without manual annotations. Experiments show \ourmethod{} generates high-quality, geometry-consistent driving videos, generalizes to diverse driving scenarios, and enhances perception and planning tasks in autonomous driving.  


%% file: root.bbl
\begin{thebibliography}{10}
\providecommand{\url}[1]{#1}
\csname url@samestyle\endcsname
\providecommand{\newblock}{\relax}
\providecommand{\bibinfo}[2]{#2}
\providecommand{\BIBentrySTDinterwordspacing}{\spaceskip=0pt\relax}
\providecommand{\BIBentryALTinterwordstretchfactor}{4}
\providecommand{\BIBentryALTinterwordspacing}{\spaceskip=\fontdimen2\font plus
\BIBentryALTinterwordstretchfactor\fontdimen3\font minus \fontdimen4\font\relax}
\providecommand{\BIBforeignlanguage}[2]{{%
\expandafter\ifx\csname l@#1\endcsname\relax
\typeout{** WARNING: IEEEtran.bst: No hyphenation pattern has been}%
\typeout{** loaded for the language `#1'. Using the pattern for}%
\typeout{** the default language instead.}%
\else
\language=\csname l@#1\endcsname
\fi
#2}}
\providecommand{\BIBdecl}{\relax}
\BIBdecl

\bibitem{nsg}
J.~Ost, F.~Mannan, N.~Thuerey, J.~Knodt, and F.~Heide, ``Neural scene graphs for dynamic scenes,'' in \emph{Proceedings of the IEEE/CVF Conference on Computer Vision and Pattern Recognition}, 2021, pp. 2856--2865.

\bibitem{unisim}
Z.~Yang, Y.~Chen, J.~Wang, S.~Manivasagam, W.-C. Ma, A.~J. Yang, and R.~Urtasun, ``Unisim: A neural closed-loop sensor simulator,'' in \emph{Proceedings of the IEEE/CVF Conference on Computer Vision and Pattern Recognition}, 2023, pp. 1389--1399.

\bibitem{emernerf}
J.~Yang, B.~Ivanovic, O.~Litany, X.~Weng, S.~W. Kim, B.~Li, T.~Che, D.~Xu, S.~Fidler, M.~Pavone \emph{et~al.}, ``Emernerf: Emergent spatial-temporal scene decomposition via self-supervision,'' \emph{arXiv preprint arXiv:2311.02077}, 2023.

\bibitem{drivinggaussian}
X.~Zhou, Z.~Lin, X.~Shan, Y.~Wang, D.~Sun, and M.-H. Yang, ``Drivinggaussian: Composite gaussian splatting for surrounding dynamic autonomous driving scenes,'' in \emph{Proceedings of the IEEE/CVF Conference on Computer Vision and Pattern Recognition}, 2024, pp. 21\,634--21\,643.

\bibitem{streetgaussian}
Y.~Yan, H.~Lin, C.~Zhou, W.~Wang, H.~Sun, K.~Zhan, X.~Lang, X.~Zhou, and S.~Peng, ``Street gaussians for modeling dynamic urban scenes,'' \emph{arXiv preprint arXiv:2401.01339}, 2024.

\bibitem{s3gaussian}
N.~Huang, X.~Wei, W.~Zheng, P.~An, M.~Lu, W.~Zhan, M.~Tomizuka, K.~Keutzer, and S.~Zhang, ``S3 gaussian: Self-supervised street gaussians for autonomous driving,'' \emph{arXiv preprint arXiv:2405.20323}, 2024.

\bibitem{pvg}
Y.~Chen, C.~Gu, J.~Jiang, X.~Zhu, and L.~Zhang, ``Periodic vibration gaussian: Dynamic urban scene reconstruction and real-time rendering,'' \emph{arXiv preprint arXiv:2311.18561}, 2023.

\bibitem{nerf}
B.~Mildenhall, P.~P. Srinivasan, M.~Tancik, J.~T. Barron, R.~Ramamoorthi, and R.~Ng, ``Nerf: Representing scenes as neural radiance fields for view synthesis,'' \emph{Communications of the ACM}, vol.~65, no.~1, pp. 99--106, 2021.

\bibitem{3dgs}
B.~Kerbl, G.~Kopanas, T.~Leimk{\"u}hler, and G.~Drettakis, ``3d gaussian splatting for real-time radiance field rendering.'' \emph{ACM Trans. Graph.}, vol.~42, no.~4, pp. 139--1, 2023.

\bibitem{gaia1}
A.~Hu, L.~Russell, H.~Yeo, Z.~Murez, G.~Fedoseev, A.~Kendall, J.~Shotton, and G.~Corrado, ``Gaia-1: A generative world model for autonomous driving,'' \emph{arXiv preprint arXiv:2309.17080}, 2023.

\bibitem{magicdrive}
R.~Gao, K.~Chen, E.~Xie, L.~Hong, Z.~Li, D.-Y. Yeung, and Q.~Xu, ``Magicdrive: Street view generation with diverse 3d geometry control,'' \emph{arXiv preprint arXiv:2310.02601}, 2023.

\bibitem{drivewm}
Y.~Wang, J.~He, L.~Fan, H.~Li, Y.~Chen, and Z.~Zhang, ``Driving into the future: Multiview visual forecasting and planning with world model for autonomous driving,'' in \emph{Proceedings of the IEEE/CVF Conference on Computer Vision and Pattern Recognition}, 2024, pp. 14\,749--14\,759.

\bibitem{drivedreamer}
X.~Wang, Z.~Zhu, G.~Huang, X.~Chen, and J.~Lu, ``Drivedreamer: Towards real-world-driven world models for autonomous driving,'' \emph{arXiv preprint arXiv:2309.09777}, 2023.

\bibitem{genad}
J.~Yang, S.~Gao, Y.~Qiu, L.~Chen, T.~Li, B.~Dai, K.~Chitta, P.~Wu, J.~Zeng, P.~Luo \emph{et~al.}, ``Generalized predictive model for autonomous driving,'' in \emph{Proceedings of the IEEE/CVF Conference on Computer Vision and Pattern Recognition}, 2024, pp. 14\,662--14\,672.

\bibitem{vista}
S.~Gao, J.~Yang, L.~Chen, K.~Chitta, Y.~Qiu, A.~Geiger, J.~Zhang, and H.~Li, ``Vista: A generalizable driving world model with high fidelity and versatile controllability,'' \emph{arXiv preprint arXiv:2405.17398}, 2024.

\bibitem{sd}
R.~Rombach, A.~Blattmann, D.~Lorenz, P.~Esser, and B.~Ommer, ``High-resolution image synthesis with latent diffusion models,'' in \emph{Proceedings of the IEEE/CVF conference on computer vision and pattern recognition}, 2022, pp. 10\,684--10\,695.

\bibitem{svd}
A.~Blattmann, T.~Dockhorn, S.~Kulal, D.~Mendelevitch, M.~Kilian, D.~Lorenz, Y.~Levi, Z.~English, V.~Voleti, A.~Letts \emph{et~al.}, ``Stable video diffusion: Scaling latent video diffusion models to large datasets,'' \emph{arXiv preprint arXiv:2311.15127}, 2023.

\bibitem{nuscenes}
H.~Caesar, V.~Bankiti, A.~H. Lang, S.~Vora, V.~E. Liong, Q.~Xu, A.~Krishnan, Y.~Pan, G.~Baldan, and O.~Beijbom, ``nuscenes: A multimodal dataset for autonomous driving,'' in \emph{Proceedings of the IEEE/CVF conference on computer vision and pattern recognition}, 2020, pp. 11\,621--11\,631.

\bibitem{waymo}
P.~Sun, H.~Kretzschmar, X.~Dotiwalla, A.~Chouard, V.~Patnaik, P.~Tsui, J.~Guo, Y.~Zhou, Y.~Chai, B.~Caine \emph{et~al.}, ``Scalability in perception for autonomous driving: Waymo open dataset,'' in \emph{Proceedings of the IEEE/CVF conference on computer vision and pattern recognition}, 2020, pp. 2446--2454.

\bibitem{deformablegs}
Z.~Yang, X.~Gao, W.~Zhou, S.~Jiao, Y.~Zhang, and X.~Jin, ``Deformable 3d gaussians for high-fidelity monocular dynamic scene reconstruction,'' in \emph{Proceedings of the IEEE/CVF Conference on Computer Vision and Pattern Recognition}, 2024, pp. 20\,331--20\,341.

\bibitem{bevgen}
A.~Swerdlow, R.~Xu, and B.~Zhou, ``Street-view image generation from a bird's-eye view layout,'' \emph{IEEE Robotics and Automation Letters}, 2024.

\bibitem{mars}
Z.~Wu, T.~Liu, L.~Luo, Z.~Zhong, J.~Chen, H.~Xiao, C.~Hou, H.~Lou, Y.~Chen, R.~Yang \emph{et~al.}, ``Mars: An instance-aware, modular and realistic simulator for autonomous driving,'' in \emph{CAAI International Conference on Artificial Intelligence}.\hskip 1em plus 0.5em minus 0.4em\relax Springer, 2023, pp. 3--15.

\bibitem{sv4d}
Y.~Xie, C.-H. Yao, V.~Voleti, H.~Jiang, and V.~Jampani, ``Sv4d: Dynamic 3d content generation with multi-frame and multi-view consistency,'' \emph{arXiv preprint arXiv:2407.17470}, 2024.

\bibitem{l4gm}
J.~Ren, K.~Xie, A.~Mirzaei, H.~Liang, X.~Zeng, K.~Kreis, Z.~Liu, A.~Torralba, S.~Fidler, S.~W. Kim \emph{et~al.}, ``L4gm: Large 4d gaussian reconstruction model,'' \emph{arXiv preprint arXiv:2406.10324}, 2024.

\bibitem{dreamscene4d}
W.-H. Chu, L.~Ke, and K.~Fragkiadaki, ``Dreamscene4d: Dynamic multi-object scene generation from monocular videos,'' \emph{arXiv preprint arXiv:2405.02280}, 2024.

\bibitem{diffusion4d}
H.~Liang, Y.~Yin, D.~Xu, H.~Liang, Z.~Wang, K.~N. Plataniotis, Y.~Zhao, and Y.~Wei, ``Diffusion4d: Fast spatial-temporal consistent 4d generation via video diffusion models,'' \emph{arXiv preprint arXiv:2405.16645}, 2024.

\bibitem{lgm}
J.~Tang, Z.~Chen, X.~Chen, T.~Wang, G.~Zeng, and Z.~Liu, ``Lgm: Large multi-view gaussian model for high-resolution 3d content creation,'' \emph{arXiv preprint arXiv:2402.05054}, 2024.

\bibitem{gaussianflow}
Q.~Gao, Q.~Xu, Z.~Cao, B.~Mildenhall, W.~Ma, L.~Chen, D.~Tang, and U.~Neumann, ``Gaussianflow: Splatting gaussian dynamics for 4d content creation,'' \emph{arXiv preprint arXiv:2403.12365}, 2024.

\bibitem{comp4d}
D.~Xu, H.~Liang, N.~P. Bhatt, H.~Hu, H.~Liang, K.~N. Plataniotis, and Z.~Wang, ``Comp4d: Llm-guided compositional 4d scene generation,'' \emph{arXiv preprint arXiv:2403.16993}, 2024.

\bibitem{gslrm}
K.~Zhang, S.~Bi, H.~Tan, Y.~Xiangli, N.~Zhao, K.~Sunkavalli, and Z.~Xu, ``Gs-lrm: Large reconstruction model for 3d gaussian splatting,'' \emph{arXiv preprint arXiv:2404.19702}, 2024.

\bibitem{4k4dgen}
R.~Li, P.~Pan, B.~Yang, D.~Xu, S.~Zhou, X.~Zhang, Z.~Li, A.~Kadambi, Z.~Wang, and Z.~Fan, ``4k4dgen: Panoramic 4d generation at 4k resolution,'' \emph{arXiv preprint arXiv:2406.13527}, 2024.

\bibitem{textto4d}
U.~Singer, S.~Sheynin, A.~Polyak, O.~Ashual, I.~Makarov, F.~Kokkinos, N.~Goyal, A.~Vedaldi, D.~Parikh, J.~Johnson \emph{et~al.}, ``Text-to-4d dynamic scene generation,'' \emph{arXiv preprint arXiv:2301.11280}, 2023.

\bibitem{luciddreamer}
J.~Chung, S.~Lee, H.~Nam, J.~Lee, and K.~M. Lee, ``Luciddreamer: Domain-free generation of 3d gaussian splatting scenes,'' \emph{arXiv preprint arXiv:2311.13384}, 2023.

\bibitem{alignyourgaussian}
H.~Ling, S.~W. Kim, A.~Torralba, S.~Fidler, and K.~Kreis, ``Align your gaussians: Text-to-4d with dynamic 3d gaussians and composed diffusion models,'' in \emph{Proceedings of the IEEE/CVF Conference on Computer Vision and Pattern Recognition}, 2024, pp. 8576--8588.

\bibitem{nfldm}
S.~W. Kim, B.~Brown, K.~Yin, K.~Kreis, K.~Schwarz, D.~Li, R.~Rombach, A.~Torralba, and S.~Fidler, ``Neuralfield-ldm: Scene generation with hierarchical latent diffusion models,'' in \emph{Proceedings of the IEEE/CVF conference on computer vision and pattern recognition}, 2023, pp. 8496--8506.

\bibitem{lrm}
Y.~Hong, K.~Zhang, J.~Gu, S.~Bi, Y.~Zhou, D.~Liu, F.~Liu, K.~Sunkavalli, T.~Bui, and H.~Tan, ``Lrm: Large reconstruction model for single image to 3d,'' \emph{arXiv preprint arXiv:2311.04400}, 2023.

\bibitem{grm}
Y.~Xu, Z.~Shi, W.~Yifan, H.~Chen, C.~Yang, S.~Peng, Y.~Shen, and G.~Wetzstein, ``Grm: Large gaussian reconstruction model for efficient 3d reconstruction and generation,'' \emph{arXiv preprint arXiv:2403.14621}, 2024.

\bibitem{realmdreamer}
J.~Shriram, A.~Trevithick, L.~Liu, and R.~Ramamoorthi, ``Realmdreamer: Text-driven 3d scene generation with inpainting and depth diffusion,'' \emph{arXiv preprint arXiv:2404.07199}, 2024.

\bibitem{one2345}
M.~Liu, C.~Xu, H.~Jin, L.~Chen, M.~Varma~T, Z.~Xu, and H.~Su, ``One-2-3-45: Any single image to 3d mesh in 45 seconds without per-shape optimization,'' \emph{Advances in Neural Information Processing Systems}, vol.~36, 2024.

\bibitem{zero123}
R.~Liu, R.~Wu, B.~Van~Hoorick, P.~Tokmakov, S.~Zakharov, and C.~Vondrick, ``Zero-1-to-3: Zero-shot one image to 3d object,'' in \emph{Proceedings of the IEEE/CVF international conference on computer vision}, 2023, pp. 9298--9309.

\bibitem{dreamfusion}
B.~Poole, A.~Jain, J.~T. Barron, and B.~Mildenhall, ``Dreamfusion: Text-to-3d using 2d diffusion,'' \emph{arXiv preprint arXiv:2209.14988}, 2022.

\bibitem{4real}
H.~Yu, C.~Wang, P.~Zhuang, W.~Menapace, A.~Siarohin, J.~Cao, L.~A. Jeni, S.~Tulyakov, and H.-Y. Lee, ``4real: Towards photorealistic 4d scene generation via video diffusion models,'' \emph{arXiv preprint arXiv:2406.07472}, 2024.

\bibitem{dreamgaussian4d}
J.~Ren, L.~Pan, J.~Tang, C.~Zhang, A.~Cao, G.~Zeng, and Z.~Liu, ``Dreamgaussian4d: Generative 4d gaussian splatting,'' \emph{arXiv preprint arXiv:2312.17142}, 2023.

\bibitem{4dfy}
S.~Bahmani, I.~Skorokhodov, V.~Rong, G.~Wetzstein, L.~Guibas, P.~Wonka, S.~Tulyakov, J.~J. Park, A.~Tagliasacchi, and D.~B. Lindell, ``4d-fy: Text-to-4d generation using hybrid score distillation sampling,'' in \emph{Proceedings of the IEEE/CVF Conference on Computer Vision and Pattern Recognition}, 2024, pp. 7996--8006.

\bibitem{diffusionprior}
C.~Wang, P.~Zhuang, A.~Siarohin, J.~Cao, G.~Qian, H.-Y. Lee, and S.~Tulyakov, ``Diffusion priors for dynamic view synthesis from monocular videos,'' \emph{arXiv preprint arXiv:2401.05583}, 2024.

\bibitem{magicdrive3d}
R.~Gao, K.~Chen, Z.~Li, L.~Hong, Z.~Li, and Q.~Xu, ``Magicdrive3d: Controllable 3d generation for any-view rendering in street scenes,'' \emph{arXiv preprint arXiv:2405.14475}, 2024.

\bibitem{colmap}
J.~L. Schonberger and J.-M. Frahm, ``Structure-from-motion revisited,'' in \emph{Proceedings of the IEEE conference on computer vision and pattern recognition}, 2016, pp. 4104--4113.

\bibitem{dust3r}
S.~Wang, V.~Leroy, Y.~Cabon, B.~Chidlovskii, and J.~Revaud, ``Dust3r: Geometric 3d vision made easy,'' in \emph{Proceedings of the IEEE/CVF Conference on Computer Vision and Pattern Recognition}, 2024, pp. 20\,697--20\,709.

\bibitem{weiszfeld}
F.~Plastria, ``The weiszfeld algorithm: proof, amendments, and extensions,'' \emph{Foundations of location analysis}, pp. 357--389, 2011.

\bibitem{dnerf}
A.~Pumarola, E.~Corona, G.~Pons-Moll, and F.~Moreno-Noguer, ``D-nerf: Neural radiance fields for dynamic scenes,'' in \emph{Proceedings of the IEEE/CVF Conference on Computer Vision and Pattern Recognition}, 2021, pp. 10\,318--10\,327.

\bibitem{d2nerf}
T.~Wu, F.~Zhong, A.~Tagliasacchi, F.~Cole, and C.~Oztireli, ``D\^{} 2nerf: Self-supervised decoupling of dynamic and static objects from a monocular video,'' \emph{Advances in neural information processing systems}, vol.~35, pp. 32\,653--32\,666, 2022.

\bibitem{nerfhugs}
J.~Chen, Y.~Qin, L.~Liu, J.~Lu, and G.~Li, ``Nerf-hugs: Improved neural radiance fields in non-static scenes using heuristics-guided segmentation,'' in \emph{Proceedings of the IEEE/CVF Conference on Computer Vision and Pattern Recognition}, 2024, pp. 19\,436--19\,446.

\bibitem{nerfonthego}
W.~Ren, Z.~Zhu, B.~Sun, J.~Chen, M.~Pollefeys, and S.~Peng, ``Nerf on-the-go: Exploiting uncertainty for distractor-free nerfs in the wild,'' in \emph{Proceedings of the IEEE/CVF Conference on Computer Vision and Pattern Recognition}, 2024, pp. 8931--8940.

\bibitem{robustnerf}
S.~Sabour, S.~Vora, D.~Duckworth, I.~Krasin, D.~J. Fleet, and A.~Tagliasacchi, ``Robustnerf: Ignoring distractors with robust losses,'' in \emph{Proceedings of the IEEE/CVF Conference on Computer Vision and Pattern Recognition}, 2023, pp. 20\,626--20\,636.

\bibitem{wildgaussians}
J.~Kulhanek, S.~Peng, Z.~Kukelova, M.~Pollefeys, and T.~Sattler, ``Wildgaussians: 3d gaussian splatting in the wild,'' \emph{arXiv preprint arXiv:2407.08447}, 2024.

\bibitem{wildgs}
J.~Xu, Y.~Mei, and V.~M. Patel, ``Wild-gs: Real-time novel view synthesis from unconstrained photo collections,'' \emph{arXiv preprint arXiv:2406.10373}, 2024.

\bibitem{spotlesssplats}
S.~Sabour, L.~Goli, G.~Kopanas, M.~Matthews, D.~Lagun, L.~Guibas, A.~Jacobson, D.~J. Fleet, and A.~Tagliasacchi, ``Spotlesssplats: Ignoring distractors in 3d gaussian splatting,'' \emph{arXiv preprint arXiv:2406.20055}, 2024.

\bibitem{ssimloss}
H.~Zhao, O.~Gallo, I.~Frosio, and J.~Kautz, ``Loss functions for image restoration with neural networks,'' \emph{IEEE Transactions on computational imaging}, vol.~3, no.~1, pp. 47--57, 2016.

\bibitem{cvt}
B.~Zhou and P.~Kr{\"a}henb{\"u}hl, ``Cross-view transformers for real-time map-view semantic segmentation,'' in \emph{Proceedings of the IEEE/CVF conference on computer vision and pattern recognition}, 2022, pp. 13\,760--13\,769.

\bibitem{bevcontrol}
K.~Yang, E.~Ma, J.~Peng, Q.~Guo, D.~Lin, and K.~Yu, ``Bevcontrol: Accurately controlling street-view elements with multi-perspective consistency via bev sketch layout,'' \emph{arXiv preprint arXiv:2308.01661}, 2023.

\bibitem{gptdriver}
J.~Mao, Y.~Qian, H.~Zhao, and Y.~Wang, ``Gpt-driver: Learning to drive with gpt,'' \emph{arXiv preprint arXiv:2310.01415}, 2023.

\bibitem{uniad}
Y.~Hu, J.~Yang, L.~Chen, K.~Li, C.~Sima, X.~Zhu, S.~Chai, S.~Du, T.~Lin, W.~Wang \emph{et~al.}, ``Planning-oriented autonomous driving,'' in \emph{Proceedings of the IEEE/CVF Conference on Computer Vision and Pattern Recognition}, 2023, pp. 17\,853--17\,862.

\bibitem{stp3}
S.~Hu, L.~Chen, P.~Wu, H.~Li, J.~Yan, and D.~Tao, ``St-p3: End-to-end vision-based autonomous driving via spatial-temporal feature learning,'' in \emph{European Conference on Computer Vision}.\hskip 1em plus 0.5em minus 0.4em\relax Springer, 2022, pp. 533--549.

\bibitem{vad}
B.~Jiang, S.~Chen, Q.~Xu, B.~Liao, J.~Chen, H.~Zhou, Q.~Zhang, W.~Liu, C.~Huang, and X.~Wang, ``Vad: Vectorized scene representation for efficient autonomous driving,'' in \emph{Proceedings of the IEEE/CVF International Conference on Computer Vision}, 2023, pp. 8340--8350.

\bibitem{nmp}
W.~Zeng, W.~Luo, S.~Suo, A.~Sadat, B.~Yang, S.~Casas, and R.~Urtasun, ``End-to-end interpretable neural motion planner,'' in \emph{Proceedings of the IEEE/CVF Conference on Computer Vision and Pattern Recognition}, 2019, pp. 8660--8669.

\bibitem{ff}
P.~Hu, A.~Huang, J.~Dolan, D.~Held, and D.~Ramanan, ``Safe local motion planning with self-supervised freespace forecasting,'' in \emph{Proceedings of the IEEE/CVF Conference on Computer Vision and Pattern Recognition}, 2021, pp. 12\,732--12\,741.

\bibitem{eo}
T.~Khurana, P.~Hu, A.~Dave, J.~Ziglar, D.~Held, and D.~Ramanan, ``Differentiable raycasting for self-supervised occupancy forecasting,'' in \emph{European Conference on Computer Vision}.\hskip 1em plus 0.5em minus 0.4em\relax Springer, 2022, pp. 353--369.

\end{thebibliography}
